# A High Performance Fingerprint Liveness Detection Method Based on Quality Related Features

Javier Galbally[a,*], Fernando Alonso-Fernandez[a], Julian Fierrez[a], Javier Ortega-Garcia[a]

[a]*Biometric Recognition Group - ATVS, EPS, Universidad Autonoma de Madrid  
C/ Francisco Tomas y Valiente, 11 - 28049 Madrid, Spain.*

**Abstract**

A new software-based liveness detection approach using a novel fingerprint parameterization based on quality related features is proposed. The system is tested on a highly challenging database comprising over 10,500 real and fake images acquired with five sensors of different technologies and covering a wide range of direct attack scenarios in terms of materials and procedures followed to generate the gummy fingers. The proposed solution proves to be robust to the multi-scenario dataset, and presents an overall rate of 90% correctly classified samples. Furthermore, the liveness detection method presented has the added advantage over previously studied techniques of needing just one image from a finger to decide whether it is real or fake. This last characteristic provides the method with very valuable features as it makes it less intrusive, more user friendly, faster and reduces its implementation costs.

## 1. Introduction

Automatic access of persons to services is becoming increasingly important in the information era. This has resulted in the establishment of a new technological area known as biometric recognition, or simply *biometrics* [1]. The basic aim of biometrics is to discriminate automatically between subjects in a reliable way and according to some target application based on one or more signals derived from physical or behavioral traits, such as fingerprint, face, iris, voice, hand, or written signature.

Biometric technology presents several advantages over classical security methods based on something that you know (PIN, Password, etc.) or something that you have (key, card, etc.). Traditional authentication systems cannot discriminate between impostors who have illegally acquired the privileges to access a system and the genuine user, and cannot satisfy negative claims of identity (i.e., I am not John Doe) [1]. Furthermore, in biometric systems there is no need for the user to remember difficult PIN codes that could be easily forgotten or to carry a key that could be lost or stolen.

However, in spite of these advantages, biometric systems present a number of drawbacks [2], including the lack of secrecy (e.g., everybody knows our face or could get our fingerprints), and the fact that a biometric trait cannot be replaced (if we forget a password we can easily generate a new one, but no new fingerprint can be generated if an impostor *steals* it). Furthermore, biometric systems are vulnerable to external attacks which could decrease their level of security [3, 4, 5], thus, it is of special relevance to understand the threats to which they are subjected and to analyze their vulnerabilities in order to prevent possible attacks and propose new countermeasures that increase their benefits for the final user.

In the last recent years important research efforts have been conducted to study the vulnerabilities of biometric systems to direct attacks to the sensor (carried out using synthetic biometric traits such as gummy fingers or high quality iris printed images) [3, 6], and indirect attacks (carried out against some of the inner modules of the system) [7, 8]. Furthermore, the interest for the analysis of security vulnerabilities has surpassed the scientific field and different standardization initiatives at international level have emerged in order to deal with the problem of security evaluation in biometric systems, such as the Common Criteria through different Supporting Documents [9], or the Biometric Evaluation Methodology [10].

Within the studied vulnerabilities, special attention has been paid to direct attacks carried out against fingerprint recognition systems [11, 12, 13]. These attacking methods consist on presenting a synthetically generated fingerprint to the sensor so that it is recognized as the legitimate user and access is granted. These attacks have the advantage over other more sophisticated attacking algorithms, such as the hill-climbing strategies [7], of not needing any in-formation about the internal working of the system (e.g., features used, template format). Furthermore, as they are carried out outside the digital domain these attacks are more difficult to be detected as the digital protection mechanisms (e.g., digital signature, watermarking) are not valid to prevent them.

*Corresponding author  
Email addresses:* javier.galbally@uam.es (Javier Galbally), fernando.alonso@uam.es (Fernando Alonso-Fernandez), julian.fierrez@uam.es (Julian Fierrez), javier.ortega@uam.es (Javier Ortega-Garcia)

Two requirements have to be fulfilled by a direct attack to be successful, 1) that the attacker retrieves by some unnoticed means the legitimate user's biometric trait, and is able to generate an artefact from it (e.g., gummy finger, iris image), and 2) that the biometric system acquires and recognizes the captured sample produced with the fake trait as that of the real user. The first of the conditions is out of the reach of biometric systems designers as there will always be someone that can think of a way of illegally recovering a certain trait. Thus, researches have focused in the design of specific countermeasures that permit biometric systems to detect fake samples and reject them, improving this way the robustness of the systems against direct attacks. Among the studied anti-spoofing approaches, special attention has been paid to those known as *liveness detection* techniques, which use different physiological properties to distinguish between real and fake traits. These methods for liveness assessment represent a challenging engineering problem as they have to satisfy certain requirements [14]: *i) non-invasive*, the technique should in no case penetrate the body or present and excessive contact with the user; *ii) user friendly*, people should not be reluctant to use it; *iii) fast*, results have to be produced in very few seconds as the user cannot be asked to interact with the sensor for a long period of time; *iv) low cost*, a wide use cannot be expected if the cost is very high; *v) performance*, it should not degrade the recognition performance of the biometric system.

In the present work, we explore the potential of quality assessment (already considered in the literature for multimodal fusion [15], or score rejection [16]), for liveness detection. Thus, a new parameterization based on quality related measures for a software-based solution in fingerprint vitality detection is proposed, and its efficiency to countermeasure direct attacks is evaluated. This novel strategy has the clear advantage over previously proposed methods of needing just one fingerprint image (i.e., the same fingerprint image used for access) to extract the necessary features in order to determine if the finger presented to the sensor is real or fake. This fact shortens the acquisition process and reduces the inconvenience for the final user, complying this way with the requirements of a liveness detection approach given above: non-invasive, user friendly, fast, and low cost (being a software-based solution it does not need of any additional hardware to be embedded in the acquisition device which would raise the price).

The performance of the proposed method is evaluated on the database provided in the Fingerprint Liveness Detection Competition LivDet 2009 [17], and on a publicly available database captured at the ATVS group [13]. The complete experimental dataset comprises over 10,500 real and fake images captured with five different sensors. It contains fake samples produced with the most popular materials used in gummy finger generation (silicone, gelatin, and playdoh), and following both a cooperative and non-cooperative process. The experimental results obtained by the proposed liveness detection approach on these challenging dataset show that it can be a very powerful tool to detect gummy fingers (almost 90% of correctly classified images), and of great utility to be included in real applications in order to prevent the different types of direct attacks which have been considered in the literature.

The paper is structured as follows. In Sect. 2 a summary of the most relevant related works to the preset study is given. The overall liveness detection method is presented in Sect. 3. In Sect. 4 the databases used in the experimental protocol are described. Results are given in Sect. 5. Conclusions are finally drawn in Sect. 6.

## 2. Related Works

Different liveness detection algorithms have been proposed for traits such as fingerprint [18, 19, 20], face [21, 22, 23], or iris [24, 25, 26]. These algorithms can broadly be divided into:

- **Software-based techniques**. In this case fake traits are detected once the sample has been acquired with a standard sensor (i.e., features used to distinguish between real and fake fingers are extracted from the fingerprint image, and not from the finger itself). These approaches include the use of skin perspiration [19], or iris texture [26]. Software-based approaches can make use of *static features* being those which require one or more impressions (e.g., the finger is placed and lifted from the sensor one or more times), or *dynamic features* which are those extracted from multiple image frames (e.g., the finger is placed on the sensor for a sort time and a video sequence is captured and analyzed).

  The liveness detection method proposed and evaluated in the present work belongs to this class of techniques.

- **Hardware-based techniques**. In this case some specific device is added to the sensor in order to detect particular properties of a living trait such as the blood pressure [27], the odor [28], or the pupil hippus [25].

Software-based techniques have the advantage over the hardware-based ones of being less expensive (as no extra device in needed), and less intrusive for the user (very important characteristic for a practical liveness detection solution) [29, 20].

For the particular case of liveness detection methods for fingerprint verification systems, different solutions have been proposed in the literature. Regarding software-based approaches, two main groups can be distinguished depending on the skin features measured: those methods based on features related to the skin perspiration, and those using skin elasticity properties. In the case of hardware-based solutions, different possibilities have been explored, including the skin odor, the heart beat, or the blood pressure.

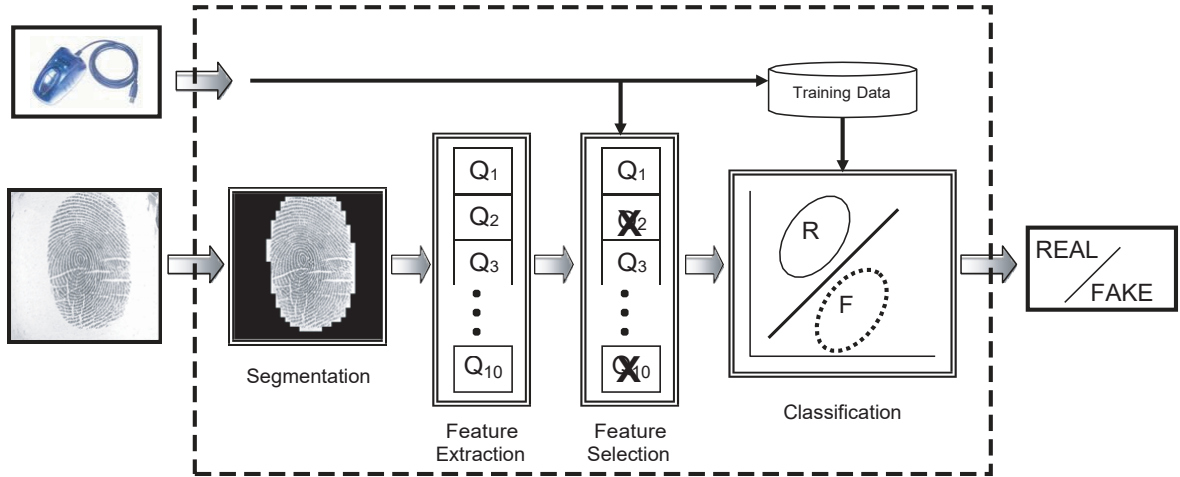

Figure 1: General diagram of the liveness detection system presented in this work.

One of the first efforts in fingerprint liveness detection was carried out by [30] who initiated a research line using the skin perspiration pattern. In this work they considered the periodicity of sweat and the sweat diffusion pattern as a way to detect fake fingerprints using a ridge signal algorithm. In a subsequent work [31], they applied a wavelet-based algorithm improving the performance reached in their initial study, and, yet in a further step [19], they extended both works with a new intensity-based perspiration liveness detection technique which leads to detection rates of around 90% on a proprietary database. Recently, a novel region-based liveness detection approach also based on perspiration features and another technique analyzing the valley noise have been proposed by the same group [32, 33].

Different fingerprint distortion models have been described in the literature [34, 35, 36], which have led to the development of liveness detection techniques based on the flexibility properties of the skin [18, 37, 38]. In particular, the liveness detection approach proposed by [37] is based on the differentiation of three fingerprint regions, namely: *i)* an inner region in direct contact with the sensor where the pressure does not allow any elastic deformation, *ii)* an external region where the pressure is very light and the skin follows the finger movements, and *iii)* an intermediate region where skin stretching and compressions take place in order to smoothly combine the previous two. In the acquisition process the user is asked to deliberately rotate his finger when removing it from the sensor surface producing this way a specific type of skin distortion which is later used as a fingerprint liveness measure. The method, which proved to be quite successful (90% detection rates of the artificial fingers are reported), was later implemented in a prototype sensor by the company Biometrika [39].

The same research group developed, in parallel to the skin elasticity method, a liveness detection procedure based on the corporal odor. [28] use a chemical sensor to discriminate the skin odor from that of other materials such as latex, silicone or gelatin. Although the system showed a remarkable performance detecting fake fingerprints made of silicone, it still showed some weakness recognizing imitations made of other materials such as gelatine, as the sensor response was very similar to that caused by human skin.

Other liveness detection approaches for fake fingerprint detection include the analysis of perspiration and elasticity related features in fingerprint image sequences [40], the use of electric properties of the skin [41], using wavelets for the analysis of the finger tip surface texture [42], the use of the power spectrum of the fingerprint image [43], or analyzing the ring patterns of the Fourier spectrum [44].

Recently, the organizers of the First Fingerprint Liveness Detection Competition (LivDet) [17], have published a comparative analysis of different software-based solutions for fingerprint liveness detection [20]. The authors study the efficiency of several approaches and give an estimation of the best performing static and dynamic features for liveness detection.

Outside the research field some companies have also proposed different methods for fingerprint liveness detection such as the ones based on ultrasounds [45, 46], on electrical measurements (some work has been done but apparently costs are too high), or light measurements ( [47] proposed a method based on temperature changes measured on an infrared image).

## 3. Liveness Detection System

The problem of liveness detection can be seen as a two-class classification problem where an input fingerprint im-

## Fingerprint Image Quality Estimation Methods

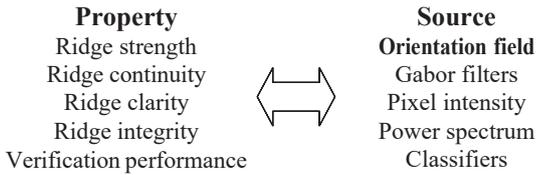

| Property | Source |
|---|---|
| Ridge strength | Orientation field |
| Ridge continuity | Gabor filters |
| Ridge clarity | Pixel intensity |
| Ridge integrity | Power spectrum |
| Verification performance | Classifiers |

Figure 2: Taxonomy of the different approaches for fingerprint image quality computation that have been described in the literature.

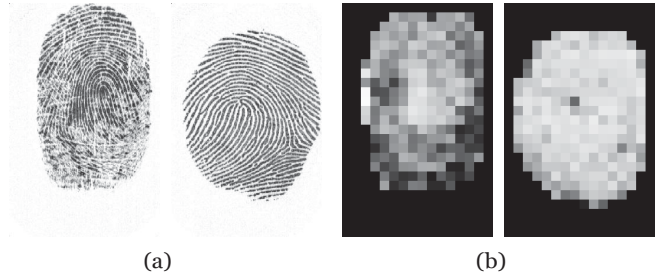

(a)            (b)

Figure 3: Computation of the Orientation Certainty Level (*OCL*) for a real and fake fingerprints. Panel (a) are the input fingerprint images (fake left, real right). Panel (b) are the block-wise values of the *OCL*; blocks with brighter color indicate higher quality in the region.

age has to be assigned to one of two classes: real or fake. The key point of the process is to find a set of discriminant features which permits to build an appropriate classifier which gives the probability of the image vitality given the extracted set of features. In the present work we propose a novel parameterization using quality measures which is tested on a complete liveness detection system.

A general diagram of the liveness detection system presented in this work is shown in Fig. 1. Two inputs are given to the system: *i)* the fingerprint image to be classified, and *ii)* the sensor used in the acquisition process.

In the first step the fingerprint is segmented from the background, for this purpose, Gabor filters are used as proposed in [48]. Once the useful information of the total image has been separated, ten different quality measures are extracted which will serve as the feature vector that will be used in the classification. Prior to the classification step, the best performing features are selected depending on the sensor that was used in the acquisition. Once the final feature vector has been generated the fingerprint is classified as real (generated by a living finger), or fake (coming from a gummy finger), using as training data of the classifier the dataset corresponding to the acquisition sensor.

### 3.1. Feature Extraction

The parameterization proposed in the present work and applied to liveness detection comprises ten quality-based features. A number of approaches for fingerprint image quality computation have been described in the literature. A taxonomy is given in [16] (see Fig. 2). Image quality can be assessed by measuring one of the following properties: ridge strength or directionality, ridge continuity, ridge clarity, integrity of the ridge-valley structure, or estimated verification performance when using the image at hand. A number of sources of information are used to measure these properties: *i)* angle information provided by the direction field, *ii)* Gabor filters, which represent another implementation of the direction angle [49], *iii)* pixel intensity of the gray-scale image, and *iv)* power spectrum. Fingerprint quality can be assessed either analyzing the image in a holistic manner, or combining the quality from local non-overlapped blocks of the image.

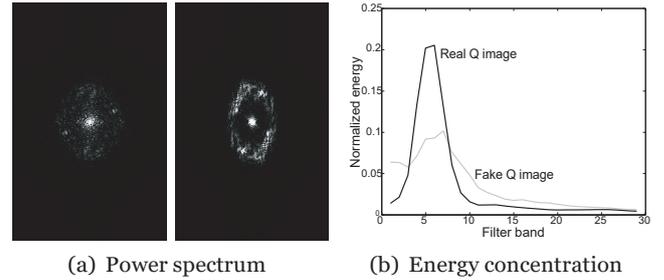

(a) Power spectrum       (b) Energy concentration

Figure 4: Computation of the energy concentration in the power spectrum for a real and fake fingerprint. Panel (a) are the power spectra of the images shown in Figure 3. Panel (b) shows the energy distributions in the region of interest. The quality values for the fake and real quality image are 0.35 and 0.88 respectively.

In the following, we give some details about the quality measures used in this paper. We have implemented several measures that make use of the above mentioned properties for quality assessment, see Table 1:

#### 3.1.1. Ridge-strength measures

- **Orientation Certainty Level ($Q_{OCL}$)** [50], which measures the energy concentration along the dominant direction of ridges using the intensity gradient. It is computed as the ratio between the two eigenvalues of the covariance matrix of the gradient vector. A relative weight is given to each region of the image based on its distance from the centroid, since regions near the centroid are supposed to provide more reliable information [51]. An example of Orientation Certainty Level computation is shown in Fig. 3 for a real and fake fingerprint samples.

- **Energy concentration in the power spectrum ($Q_E$)** [51], which is computed using ring-shaped bands. For this purpose, a set of bandpass filters is employed to extract the energy in each frequency band. High quality images will have the energy concentrated in few bands while poor ones will have a more diffused distribution. The energy concentration is measured using the entropy. An example of quality estimation using the global quality index $Q_{ENERGY}$ is shown in Fig. 4 for a real and fake fingerprint samples.

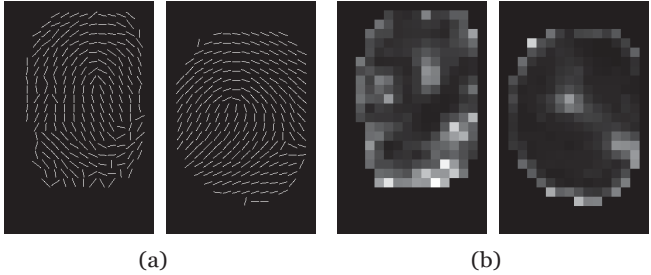

Figure 5: Computation of the Local Orientation Quality (*LOQ*) for a real and fake fingerprint samples. Panel (a) are the direction fields of the images shown in Figure 3a. Panel (b) are the block-wise values of the average absolute difference of local orientation with the surrounding blocks; blocks with brighter color indicate higher difference value and thus, lower quality.

#### 3.1.2. Ridge-continuity measures

- **Local Orientation Quality ($Q_{LOQ}$)** [52], which is computed as the average absolute difference of direction angle with the surrounding image blocks, providing information about how smoothly direction angle changes from block to block. Quality of the whole image is finally computed by averaging all the Local Orientation Quality scores of the image. In high quality images, it is expected that ridge direction changes smoothly across the whole image. An example of Local Orientation Quality computation is shown in Fig. 5 for a real and fake fingerprint samples.

- **Continuity of the orientation field ($Q_{COF}$)** [50]. This method relies on the fact that, in good quality images, ridges and valleys must flow sharply and smoothly in a locally constant direction. The direction change along rows and columns of the image is examined. Abrupt direction changes between consecutive blocks are then accumulated and mapped into a quality score. As we can observe in Fig. 5, ridge direction changes smoothly across the whole image in case of the real sample.

#### 3.1.3. Ridge-clarity measures

- **Mean ($Q_{MEAN}$)** and **standard deviation ($Q_{STD}$)** values of the gray level image, computed from the segmented foreground only. These two features had already been considered for liveness detection in [20].

- **Local Clarity Score ($Q_{LCS1}$ and $Q_{LCS2}$)** [52]. The sinusoidal-shaped wave that models ridges and valleys [53] is used to segment ridge and valley regions (see Figure 6). The clarity is then defined as the overlapping area of the gray level distributions of segmented ridges and valleys. For ridges/valleys with high clarity, both distributions should have a very small overlapping area. An example of quality estimation using the Local Clarity Score is shown in Fig. 7 for two fingerprint blocks one coming from a real fingerprint image and the other from a fake sample. It should be noted that sometimes the sinusoidal-shaped wave cannot be extracted reliably, specially in bad quality regions of the image. The quality measure $Q_{LCS1}$ discards these regions, therefore being an optimistic measure of quality. This is compensated with $Q_{LCS2}$, which does not discard these regions, but they are assigned the lowest quality level.

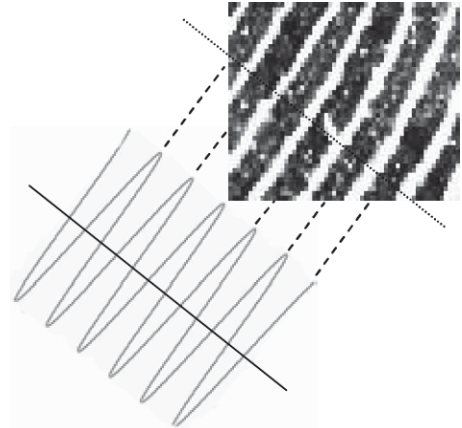

Figure 6: Modeling of ridges and valleys as a sinusoid.

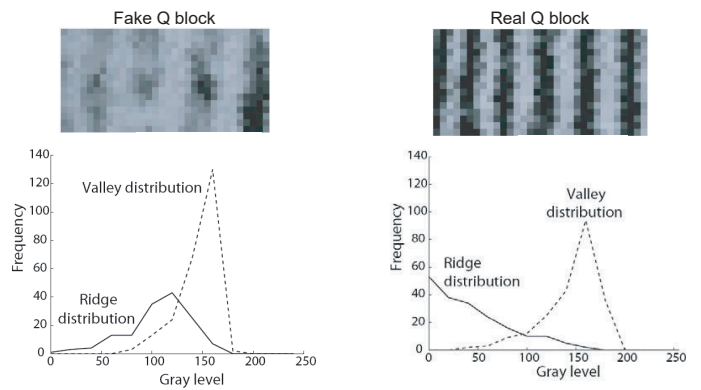

Figure 7: Computation of the Local Clarity Score for two fingerprint blocks coming from a real and fake fingerprint respectively. The fingerprint blocks appear on top, while below we show the gray level distributions of the segmented ridges and valleys. The degree of overlapping for the fake and real quality block is 0.22 and 0.10, respectively.

- **Amplitude and variance of the sinusoid that models ridges and valleys ($Q_A$ and $Q_{VAR}$)** [53]. Based on these parameters, blocks are classified as *good* and *bad*. The quality of the fingerprint is then computed as the percentage of foreground blocks marked as *good*.

### 3.2. Feature Selection

Due to the curse of dimensionality, it is possible that the best classifying results are not obtained using the set of ten proposed features, but a subset of them. As we are dealing with a ten dimensional problem there are $2^{10}-1 = 1,023$ possible feature subsets, which is a reasonably low number to apply exhaustive search as feature selection technique in order to find the best performing feature subset. This way we guarantee that we find the optimal set of features out of all the possible ones. The feature selection depends on the acquisition device (as shown in Fig. 1), as the optimal feature subsets might be different for different sensors.

| Quality measure | Property measured | Source |
|---|---|---|
| $Q_{OCL}$ | Ridge strength | Local angle |
| $Q_E$ | Ridge strength | Power spectrum |
| $Q_{LOQ}$ | Ridge continuity | Local angle |
| $Q_{COF}$ | Ridge continuity | Local angle |
| $Q_{MEAN}$ | Ridge clarity | Pixel intensity |
| $Q_{STD}$ | Ridge clarity | Pixel intensity |
| $Q_{LCS1}$ | Ridge clarity | Pixel intensity |
| $Q_{LCS2}$ | Ridge clarity | Pixel intensity |
| $Q_A$ | Ridge clarity | Pixel intensity |
| $Q_{VAR}$ | Ridge clarity | Pixel intensity |

Table 1: Summary of the quality measures used in the parameterization applied to fingerprint liveness detection.

*3.3. Classifier*

We have used Linear Discriminant Analysis (LDA) as classifier [54]. In the experiments two separate sets have been used: *i)* one for development in order to select the best performing features and to fit the two normal distributions representing each of the classes (real or fake), and *ii)* the second for test in order to evaluate the performance of the algorithm.

## 4. Datasets and Experimental Protocol

The performance of the proposed liveness detection scheme is validated on two different databases: *i)* the publicly available dataset provided in the Fingerprint Liveness Detection Competition, LivDet 2009 [17] (http://prag.diee.unica.it/LivDet09/) comprising over 18,000 real and fake samples, and *ii)* a dataset captured at the Biometric Recognition Group - ATVS [13], available at http://atvs.ii.uam.es/, which contains over 3,000 real and fake fingerprint images. Each of the databases is divided into a development set where the best feature subsets selected and the system is trained, and a test set in which the evaluation results are obtained. In order to generate totally unbiased results, there is no overlap between development and test sets (i.e., samples corresponding to each user are just included in one of the sets).

- **LivDet Database** [17]. It comprises three datasets of real and fake fingerprints captured each of them with a different flat optical sensor: *i)* Biometrika FX2000 (569 dpi), *ii)* CrossMatch Verifier 300CL (500 dpi), and *iii)* Identix DFR2100 (686dpi). The gummy fingers were generated using three different materials: silicone, gelatine and playdoh, and always with a following a consensual procedure (with the cooperation of the user). The development and test sets of this database are the same as the ones used in the LivDet competition, and their general distribution of the fingerprint images between both sets is given in Table 2. Some typical examples of the images that can be found in this database are shown in Fig. 8, where the material used for the generation of the fake fingers is given (silicone, gelatine or playdoh).

- **ATVS Database** [13]. It comprises three datasets of real and fake fingerprints captured each of them with an acquisition device of different technologies: *i)* flat optical Biometrika FX2000 (569 dpi), *ii)* flat capacitive Precise SC100 (500 dpi), and *iii)* thermal sweeping Yubee with Atmel's Fingerchip (500dpi). All the gummy fingers were generated using modeling silicone, but two different procedures were followed: with and without the cooperation of the user. Both the development and the test set contain half of the fingerprint images, and their general structure is given in Table 3. Some typical examples of the images that can be found in this database are shown in Fig. 9, where the type of process used for the generation of the gummy fingers is given (cooperative or non-cooperative).

From the summary of the two databases given in Tables 2 and 3 we can see that the complete final validation dataset comprises over 10,500 real and fake samples, divided into six different datasets (three corresponding to the LivDet DB and the other three to the ATVS DB), and captured under totally different scenarios in terms of: *i)* acquisition devices, *ii)* material used to generate the gummy fingers, and *iii)* process followed to obtain the fake images.

As was presented in Sect. 3 (and is shown in Fig 1) the proposed liveness detection system only presents two inputs: *i)* the fingerprint image to be classified, and *ii)* the sensor used to acquire that image. This way, although the material with which the different fake fingers are made is known, this fact is not used in anyway by the liveness detection system as in a real attack this information would not be available to the application. The feature selection is just made in terms of the sensor used in the acquisition.

It can be noticed from the examples shown in Figs. 8 and 9 the difficulty of the classification problem, as even for a human expert would not be easy to distinguish between the real and fake samples present at the final dataset.

## 5. Results

The performance of the proposed approach is estimated in terms of the Average Classification Error (ACE) which is defined as $ACE = (FLR + FFR)/2$, where the FLR (False Living Rate) represents the percentage of fake fingerprints misclassified as real, and the FFR (False Fake Rate) computes the percentage of real fingerprints assigned to the fake class. The evaluation scheme followed in the experimental protocol presents two successive stages, training and validation, designed to obtain totally unbiased results:

- **Stage 1: Training**. The best feature subsets are computed on the development sets defined in Tables 2 and 3. The results of this step are presented in Sect. 5.1. These optimal subsets will be the same for all the successive steps of the performance evaluation process.

- **Stage 2: Validation**. In order to obtain a better estimation of the classification capabilities of the algorithm, cross validation is performed exchanging development and test sets. The final Average Classification Error is computed as the mean of the ACE corresponding to each of the two stages of the cross validation process. Thus, the steps carried out in this stage are:

    - Step 2.1: First part of the cross validation. With the best feature subsets found in the training stage the performance of the system is computed using the development sets for training the classifier and the test sets for evaluating the performance. The result of this step is $ACE_1$ (respectively $FLR_1$ and $FRR_1$).

    - Step 2.2: Second part of the cross validation. Using the best feature subsets found in the training stage the performance of the system is computed exchanging development and test sets, that is, the test sets are used for training the classifier and the development sets for evaluating the performance. The result of this step is $ACE_2$ (respectively $FLR_2$ and $FFR_2$).

    - Step 2.3: The final classification error of the system is computed as: $ACE = (ACE_1 + ACE_2)/2$.

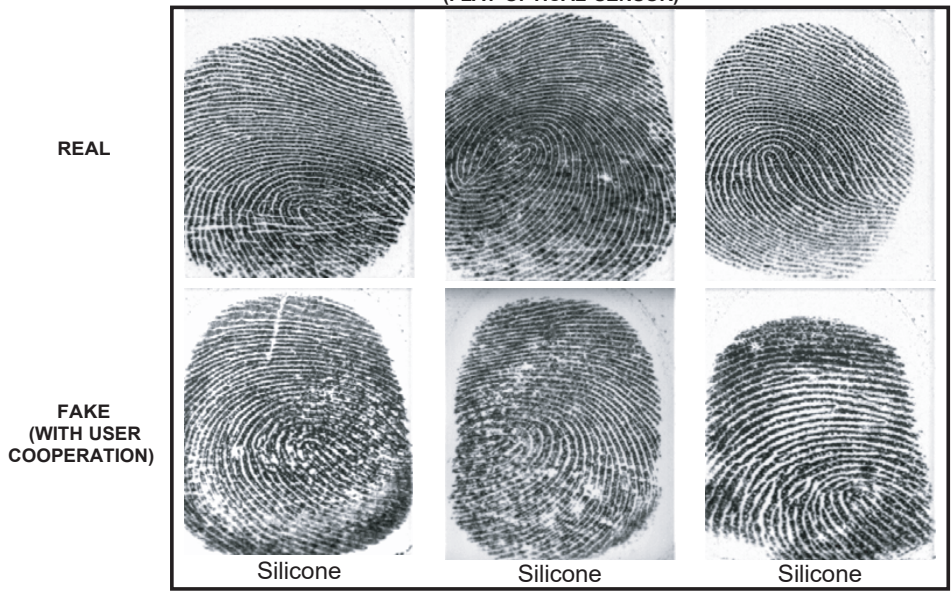
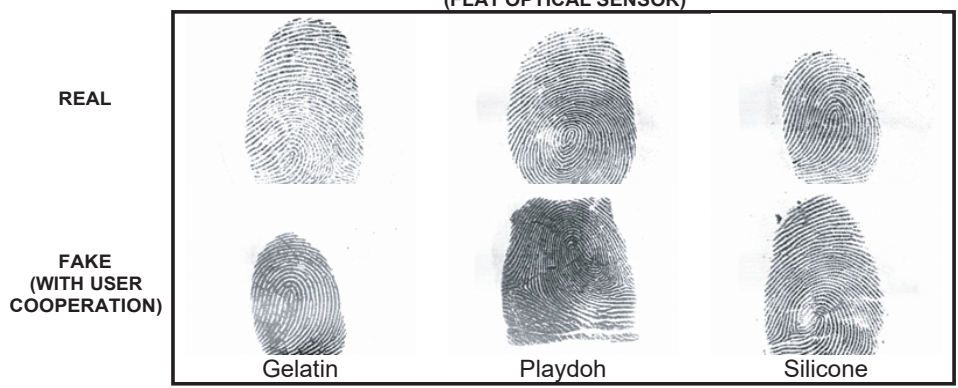
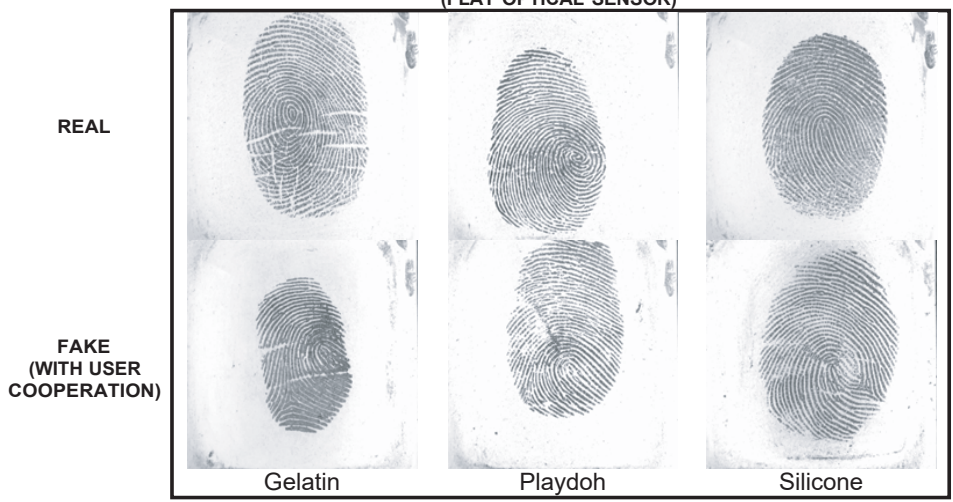

Figure 8: Typical examples of real and fake fingerprint images that can be found in the public LivDet database used in the experiments, which can be downloaded from http://prag.diee.unica.it/LivDet09/.

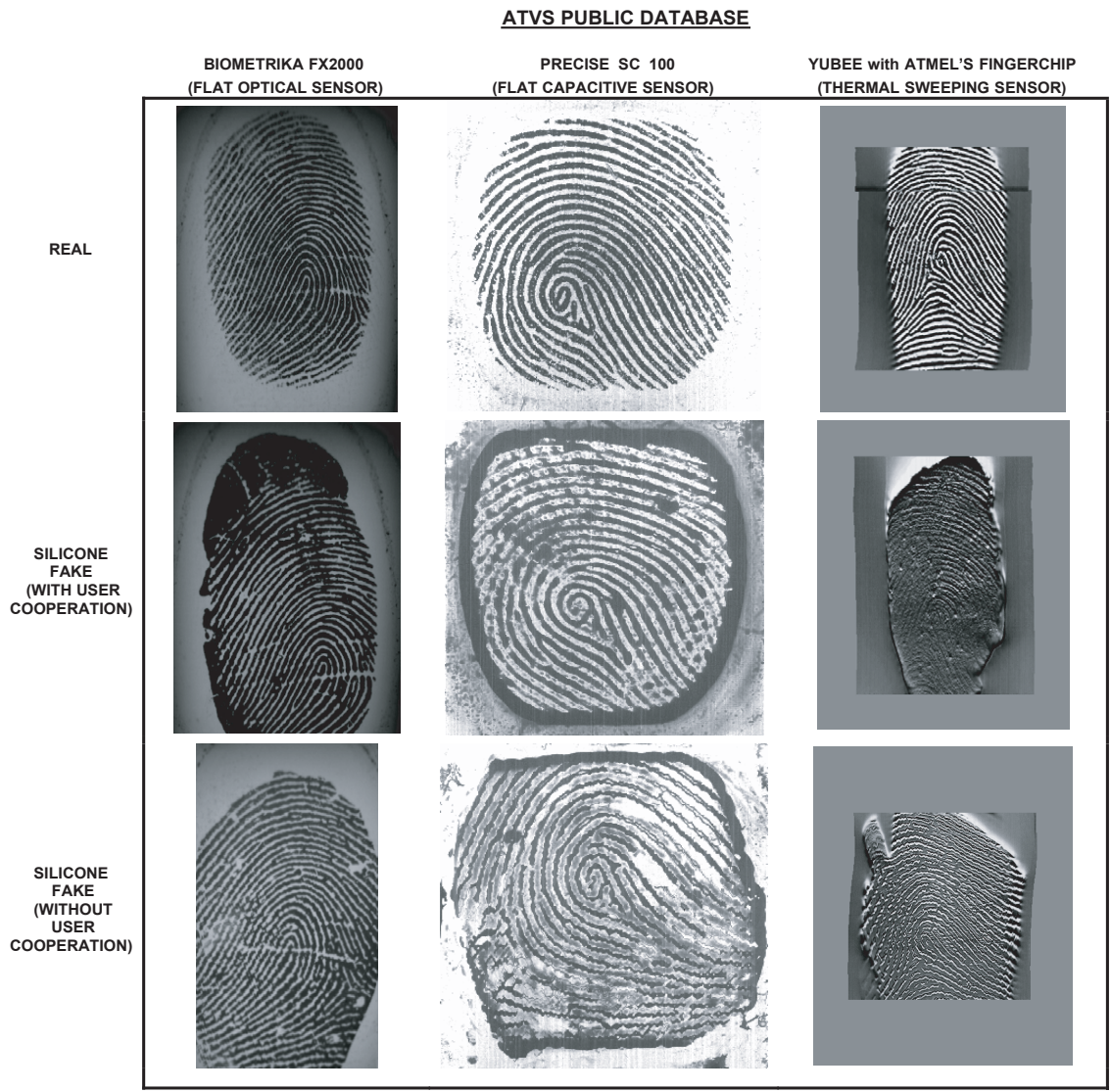

Figure 9: Typical examples of real and fake fingerprint images that can be found in the public ATVS database used in the experiments, which can be downloaded from http://atvs.ii.uam.es/.

|  | **LivDet DB** | |
| --- | --- | --- |
|  | Development (Real/Fake) | Test (Real/Fake) |
| Biometrika FX2000 (569 dpi) | 520/520s | 1473/1480s |
| CrossMatch Verifier 300CL (500 dpi) | 1000/1000 (310s+344g+346p) | 3000/3000 (930s+1036g+1034p) |
| Identix DFR2100 (686 dpi) | 750/750 (250s+250g+250p) | 2250/2250 (750s+750g+750p) |

Table 2: General structure of the LivDet DB used in the experiments. The distribution of the fake images is given in terms of the materials used for their generation: *s* for silicone, *g* for gelatin, and *p* for playdoh.

|  | **ATVS DB** | |
| --- | --- | --- |
|  | Development (Real/Fake) | Test (Real/Fake) |
| Biometrika FX2000 (569 dpi) | 255/255 (127c+128nc) | 255/255 (127c+128nc) |
| Precise SC100 (500 dpi) | 255/255 (127c+128nc) | 255/255 (127c+128nc) |
| Yubee (500 dpi) | 255/255 (127c+128nc) | 255/255 (127c+128nc) |

Table 3: General structure of the ATVS DB used in the experiments. The distribution of the fake images is given in terms of the procedure used for their generation: cooperative (*c*), or non-cooperative (*nc*).

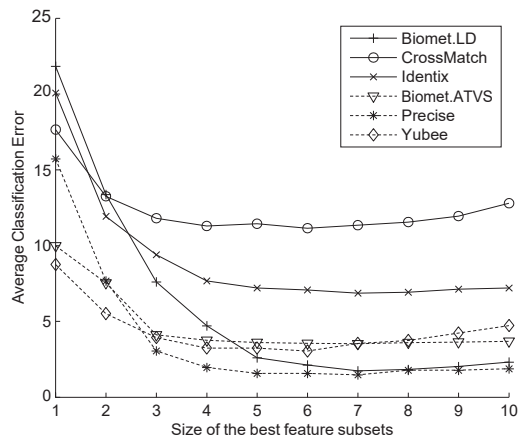

Figure 10: Evolution of the ACE for the best feature subsets with an increasing number of features, and for the three datasets.

The results obtained in each of the two stages are analyzed in the following sections.

*5.1. Stage 1: Training*

The first objective of the experiments is to find the optimal feature subsets (out of the proposed 10 feature set) for each of the different acquisition devices comprised in the two databases. Then the classification performance of each of the optimal subsets is computed on each of the datasets in terms of the Average Classification Error.

In order to find the optimal feature subsets, for each of the six datasets in the final database, the classification performance of each of the 1,023 possible feature subsets was computed on the development sets given in Tables 2 and 3 using the leave-one-out technique (i.e., all the samples in the development set are used to train the classifier except the one being classified).

| Database | Sensor | Best feature subsets for quality-based liveness detection | | | | | | | | | | ACE |
|---|---|---|---|---|---|---|---|---|---|---|---|---|
| | | Ridge Strength | | Ridge Continuity | | Ridge Clarity | | | | | | |
| | | $Q_{OCL}$ | $Q_E$ | $Q_{LOQ}$ | $Q_{COF}$ | $Q_{MEAN}$ | $Q_{STD}$ | $Q_{LCS1}$ | $Q_{LCS2}$ | $Q_A$ | $Q_{VAR}$ | |
| LivDet | Biometrika | × | × | × | | | × | × | | × | × | 1.73 |
| | CrossMatch | × | × | × | × | × | | × | | | | 11.15 |
| | Identix | × | × | × | × | × | × | | | × | | 6.87 |
| ATVS | Biometrika | × | × | × | | | × | × | | × | × | 3.53 |
| | Precise | × | × | × | | × | × | | × | | | 1.47 |
| | Yubee | × | × | | | × | × | | | × | × | 3.05 |

Table 4: Best performing feature subsets for the different datasets in the two validation databases (LivDet and ATVS). The ACE is given in %. The symbol × means that the feature is considered in the subset.

The evolution of the ACE produced by each of the best feature subsets, for an increasing number of features, and for the six datasets is shown in Fig. 10. We can observe in Fig. 10 the curse of dimensionality effect as the minimum error rate is reached in all cases for the best subset comprising 6 or 7 parameters, increasing slightly when new features are considered.

The best feature subsets found for each of the sensors are shown in Table 4, where a × means that the feature is included in the subset. The Average Classification Error for each of the best subsets is shown on the right in percentage.

From the results shown in Table 4 we can observe that there is no feature that is not included at least in one of the optimal subsets which indicates that all the proposed features are relevant for fingerprint liveness detection. Parameters $Q_E$ and $Q_{OCL}$ are present in the best feature subsets of all datasets, thus, we may conclude that the most discriminant features are those measuring the ridge strength. Also, one ridge clarity features, $Q_{STD}$, and one ridge continuity parameter $Q_{LOQ}$, are shown to provide good discriminative capabilities with all sensors (are just discarded in one of the best six subsets). On the other hand, the least useful features for liveness detection appear to be the ridge continuity related $Q_{COF}$, together with most of the rest ridge clarity features ($Q_{LCS1}$, $Q_{LCS2}$, and $Q_{VAR}$), which are only included in half or less of the best feature subsets. The information extracted from Table 4 on the discriminant capabilities of the different parameters according to the ridge property measured is summarized in Table 5.

In Table 4 we can also see that the best parameterization found for both datasets captured with the Biometrika FX2000 sensor (one from the LivDet DB and the other from the ATVS DB) is the same. This fact suggests that the best feature subsets are consistent between sets of data captured under different scenarios as long as the same acquisition device is deployed.

| Discriminative Power | | |
|---|---|---|
| Ridge Strength | Ridge Continuity | Ridge Clarity |
| High ($Q_E$, $Q_{OCL}$) | Medium ($Q_{LOQ}$) | Medium ($Q_{STD}$, $Q_{MEAN}$) |

Table 5: Summary for the six datasets of the parameters discriminant power according to the ridge property measured. The best performing features are specified in each case.

### 5.2. Stage 2: Validation

The best feature subsets found on the development sets and analyzed in Sect. 5.1, are used here to evaluate the performance of the proposed liveness detection system on the different datasets comprised in the whole evaluation database, following the general cross validation approach described in Sect. 5. The performance results of the proposed liveness detection scheme are given in Table 6 where we can see that the overall classification error of the system is around 10%.

The ACE results for the two cross validation steps are very consistent through all the different datasets except for the LivDet DB dataset captured with the Biometrika sensor (highlighted in grey). In this particular case we can observe a very big difference in the performance of the system when the training and test sets are exchanged, with an abnormally high False Fake Rate (over 50%) when the development set is used for training and the test set for evaluation. This same behaviour was observed in the rest of the participants in the LivDet competition (results of the competition are analyzed in [17]).

Furthermore, for this dataset (Biometrika.LD) there is a huge difference in the performance of the algorithm between the training stage (ACE=1.73% as is shown in the first row of Table 4) and the validation tests (ACE=26.5%); gap which is not observed in the rest of the datasets, where the classification error in both cases, training and validation, are very similar.

These facts, combined with the homogeneous performance results obtained in the rest of the datasets, suggests that there exists some inconsistency between the Biometrika development and test data provided in the LivDet competition. Most likely, the test set presents a much higher variability than the training set, producing this way very poor results when it is used for testing (in this scenario the system is not properly trained). On the
10

|  | **Performance Results in %** | | | |
| --- | --- | --- | --- | --- |
|  | FLR$_1$/FLR$_2$ | FFR$_1$/FFR$_2$ | ACE$_1$/ACE$_2$ | **ACE** |
| Biomet.LD | 3.1/9.8 | 71.8/21.5 | 37.4/15.6 | **26.5** |
| CrossMatch | 8.8/16.7 | 20.8/6.8 | 14.8/11.7 | **13.2** |
| Identix | 4.8/8.0 | 5.0/9.1 | 4.9/8.5 | **6.7** |
| Biomet.ATVS | 9.4/0.4 | 1.5/11.8 | 5.5/6.1 | **5.8** |
| Precise | 3.1/0.4 | 13.7/0.4 | 8.4/0.4 | **4.4** |
| Yubee | 2.7/1.6 | 6.3/13.0 | 4.5/7.3 | **5.9** |
| **Total** | 5.3/6.1 | 19.8/10.4 | 12.5/8.2 | **10.4** |

Table 6: Performance results of the proposed liveness detection method on the different datasets considered in the validation experiments. The subindexes 1 and 2 stand respectively for the results obtained on the first and second stages of the cross validation process.

|  | **ACE (%)** | | |
| --- | --- | --- | --- |
|  | Biomet.ATVS | Precise | Yubee |
| Cooperative | 7.0 | 6.1 | 7.6 |
| Non-Cooperative | 4.6 | 2.7 | 4.2 |

Table 7: Performance of the proposed quality-based vitality detection scheme against gummy fingers generated from a latent fingerprint following a cooperative and non-cooperative process. The datasets correspond to those comprised in the ATVS DB.

other hand, when we exchange the two sets, the training is fairly good and the classification error is much lower than in the previous case.

Thus, it may be concluded that a more realistic ACE for the Biometrika.LD dataset would be around 10%, which would also be closer to the error rate achieved in the other dataset captured with the Biometrika sensor (Biometrika.ATVS). Assuming this more realistic ACE of 10% in the case of Biometrika.LD, the final overall classification error of the proposed liveness detection scheme would be around 7.5%.

*5.2.1. Cooperative vs Non-Cooperative Attacks*

During the validation experiments, in addition to evaluating the general performance of the proposed liveness detection scheme, we also compared its potential to detect direct attacks using gummy fingers generated with and without the cooperation of the user. That is, we compared the chances of these two types of direct attacks to break a fingerprint verification system which uses the novel quality based liveness detection algorithm.

For these experiments, the feature subsets used were the ones set in the training stage for each of the three datasets comprised in the ATVS DB. It is important to notice that no specific subsets for cooperative and non-cooperative fake images were selected, as in a realistic scenario we would not know with what type of gummy finger the attacker would try to break the system. Thus, for each of the two stages of the cross validation process, the training set was the same reported in the previous experiments (containing both fake cooperative and non-cooperative samples), while the test dataset was separated into two subsets comprising one of them images coming from gummy fingers generated with the cooperation of the user, and the other without his cooperation. The ACE of the liveness detection system for the two cases is given in Table 7. In Fig. 11 we show the quality distributions of three of the considered parameters (each measuring one different fingerprint image property: ridge strength, continuity and clarity) of the different images present at the ATVS DB (genuine, fake with cooperation, and fake without cooperation), captured with the optical, capacitive and thermal sweeping sensors.

It can be observed from the results presented in Table 7 that non-cooperative attacks are easier to detect for the proposed liveness detection system than the ones with a participative user (the ACE in the first case is lower for all the studied cases than in the second). This fact could have been expected, as the non-cooperative process to generate the synthetic fingerprints is more difficult and requires more steps than the one with cooperation from the user [13], which generates further undesired effects on the final gummy finger and makes it differ more from the real trait. This effect is patent in Fig. 11 where the quality distributions of non-cooperative images are in all cases further away from the real distributions than those of the cooperative samples.

## 6. Conclusions

A novel fingerprint parameterization for liveness detection based on quality related measures has been proposed. The feature set has been used in a complete liveness detection system, and tested on two publicly available databases: *i)* the database used in the 2009 LivDet competition [17], and *ii)* a database captured at the ATVS group [13]. These two challenging databases permit to test the proposed liveness detection scheme under totally different operational scenarios in terms of the technology used by the acquisition devices (flat optical, flat capacitive, and sweeping thermal), material with which the gummy fingers are produced (gelatin, silicone and playdoh), and procedure followed to generate the fake fingers (with and without the cooperation of the user).

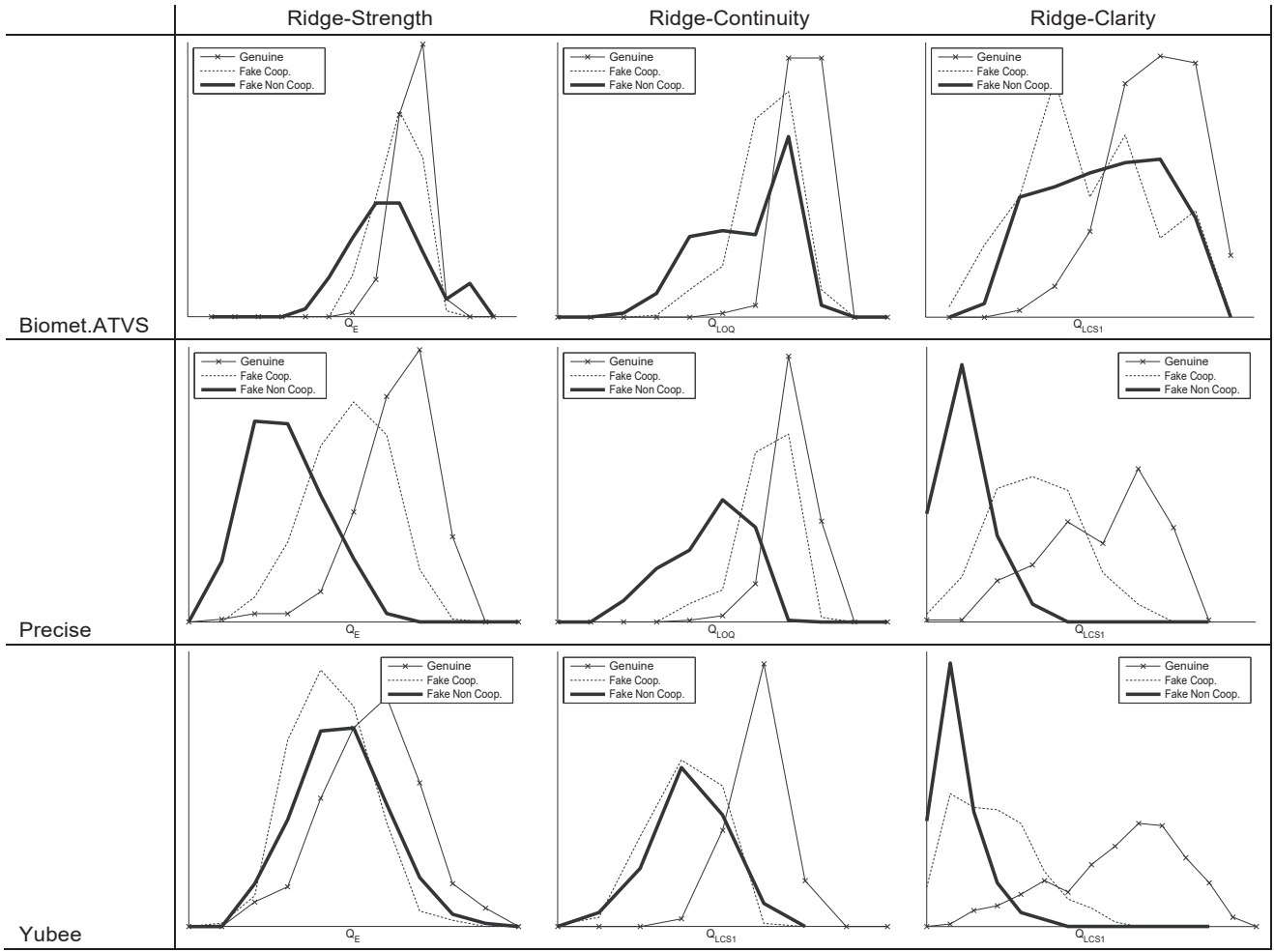

Figure 11: Quality distributions for three of the considered parameters (each measuring one different fingerprint image property) of the datasets comprised in the ATVS DB (genuine, fake with cooperation, and fake without cooperation), captured with the optical sensor, capacitive sensor, and thermal sweeping sensor.

The high performance shown by the proposed system under these completely diverse testing scenarios, correctly classifying almost 90% of the fingerprint images, proves its ability to adapt to all type of direct attacks and its efficiency as a method to minimize their effect and enhance the general security capability of fingerprint verification systems. Furthermore, using two public datasets will permit to fairly compare the results with other liveness detection techniques from the state of the art, giving an added value to the conclusions and observations extracted from the experiments

The proposed approach is part of the software-based solutions as it distinguishes between images produced by real and fake fingers based only on the acquired sample, and not on other physiological measures (e.g., odor, heartbeat, skin impedance) captured by special hardware devices added to the sensor (i.e., hardware-based solutions that increase the cost of the sensors, and are more intrusive to the user). Unlike previously presented methods, the proposed technique classifies each image in terms of features extracted from just that image, and not from different samples of the fingerprint. This way the acquisition process is faster and more convenient to the final user (that does not need to keep his finger on the sensor for a few seconds, or place it several times).

Liveness detection solutions such as the one presented in this work are of great importance in the biometric field as they help to prevent direct attacks (those carried out with synthetic traits, and very difficult to detect), enhancing this way the level of security offered to the user.

# 7. Acknowledgments

J. G. is supported by a FPU Fellowship from the Spanish MEC, F. A.-F. is supported by a Juan de la Cierva Fellowship from the Spanish MICINN. This work was supported by projects Contexts (S2009/TIC-1485) from CAM, Bio-Challenge (TEC2009-11186) from Spanish MICINN, and *Cátedra UAM-Telefónica*. The authors would also like to thank the *Dirección General de la Guardia Civil* for their support to the work.